\documentclass[10pt,twocolumn,letterpaper]{article}

\usepackage{cvpr}              

\usepackage{graphicx}
\usepackage{amsmath}
\usepackage{amssymb}
\usepackage{booktabs}
\usepackage{algorithm}
\usepackage{algpseudocode}
\usepackage{float}
\usepackage{stfloats} 
\usepackage[utf8]{inputenc} 
\usepackage[T1]{fontenc}    
\usepackage{url}            
\usepackage{booktabs}       
\usepackage{amsfonts}       
\usepackage{nicefrac}       
\usepackage{microtype}      
\usepackage{acronym}
\usepackage{graphicx}
\usepackage{wrapfig}

\usepackage{amsmath}
\usepackage{amssymb}
\usepackage{mathtools}
\usepackage{amsthm}
\usepackage{algorithm}
\usepackage{algpseudocode}
\usepackage{setspace} 
\usepackage{enumitem}
\usepackage{acronym}
\usepackage{xspace}
\usepackage[table]{xcolor}
\usepackage{graphicx}    
\usepackage{array}       
\usepackage[misc]{ifsym} 
\usepackage{makecell}    
\usepackage{booktabs}    
\usepackage{setspace}    
\everydisplay{\small} 

\usepackage{multirow}
\usepackage{indentfirst}
\usepackage{subcaption}
\usepackage{pifont}
\usepackage{lineno}
\usepackage[pagebackref,breaklinks,colorlinks]{hyperref}

\usepackage[capitalize]{cleveref}
\crefname{section}{Sec.}{Secs.}
\Crefname{section}{Section}{Sections}
\Crefname{table}{Table}{Tables}
\crefname{table}{Tab.}{Tabs.}

\def\cvprPaperID{*****} 
\def\confName{CVPR}
\def\confYear{2025}

\begin{document}

\title{IRL-VLA: Training an Vision-Language-Action Policy via Reward World Model for End-to-End Autonomous Driving}

\author{
    Anqing Jiang\textsuperscript{1*} \and
    Yu Gao\textsuperscript{1*} \and
    Yiru Wang\textsuperscript{1} \and
    Zhigang Sun\textsuperscript{1} \and
    Shuo Wang\textsuperscript{1} \and
    Yuwen Heng\textsuperscript{1} \and
    Hao Sun\textsuperscript{1\textdagger} \and 
    Shichen Tang\textsuperscript{1} \and
    Lijuan Zhu\textsuperscript{1} \and
    Jinhao Chai\textsuperscript{2} \and 
    Jijun Wang\textsuperscript{5} \and
    Zichong Gu\textsuperscript{2} \and 
    Hao Jiang\textsuperscript{3} \and 
    Li Sun\textsuperscript{4} \\
    \vspace{-0.4cm}
    \\
    \footnotesize
    \begin{tabular}{c}
        \textsuperscript{1}Bosch Corporate Research, Shanghai, China \\
        \textsuperscript{2}School of Communication and Information Engineering, Shanghai University \\
        \textsuperscript{3}School of Mechanical Engineering, Shanghai Jiao Tong University \\
        \textsuperscript{4}Bosch Mobility Solutions, Robert Bosch GmbH, Suzhou \\
        \textsuperscript{5}AIR, Tsinghua University, Beijing \\
        \vspace{0.1cm}
        *\,Equal contribution \quad \textdagger\,Corresponding author: \href{mailto:hao.sun@cn.bosch.com}{hao.sun4@cn.bosch.com}
    \end{tabular}
}
    \vspace{-0.5cm}


\maketitle

\begin{abstract}

Vision-Language-Action (VLA) models have demonstrated potential in autonomous driving. However, two critical challenges hinder their development: (1) Existing VLA architectures are typically based on imitation learning in open-loop setup which tends to capture the recorded behaviors in the dataset, leading to suboptimal and constrained  performance, (2) Close-loop training relies heavily on high-fidelity sensor simulation, where domain gaps and computational inefficiencies pose significant barriers. In this paper, we introduce IRL-VLA, a novel close-loop Reinforcement Learning via \textbf{I}nverse \textbf{R}einforcement \textbf{L}earning reward world model with a self-built VLA approach. Our framework proceeds in a three-stage paradigm: In the first stage, we propose a VLA architecture and pretrain the VLA policy via imitation learning. In the second stage, we construct a lightweight reward world model via inverse reinforcement learning to enable efficient close-loop reward computation. To further enhance planning performance, finally, we design specialized reward world model guidence reinforcement learning via PPO(Proximal Policy Optimization) to effectively balance the safety incidents, comfortable driving, and traffic efficiency. Our approach achieves state-of-the-art performance in NAVSIM v2 end-to-end driving benchmark, 1st runner up in CVPR2025 Autonomous Grand Challenge. We hope that our framework will accelerate VLA research in close-loop autonomous driving. See our project repository for more results: \url{https://github.com/IRL-VLA/IRL-VLA}

\end{abstract}

\begin{figure}[t] 
    \centering
    \includegraphics[width=0.45\textwidth]{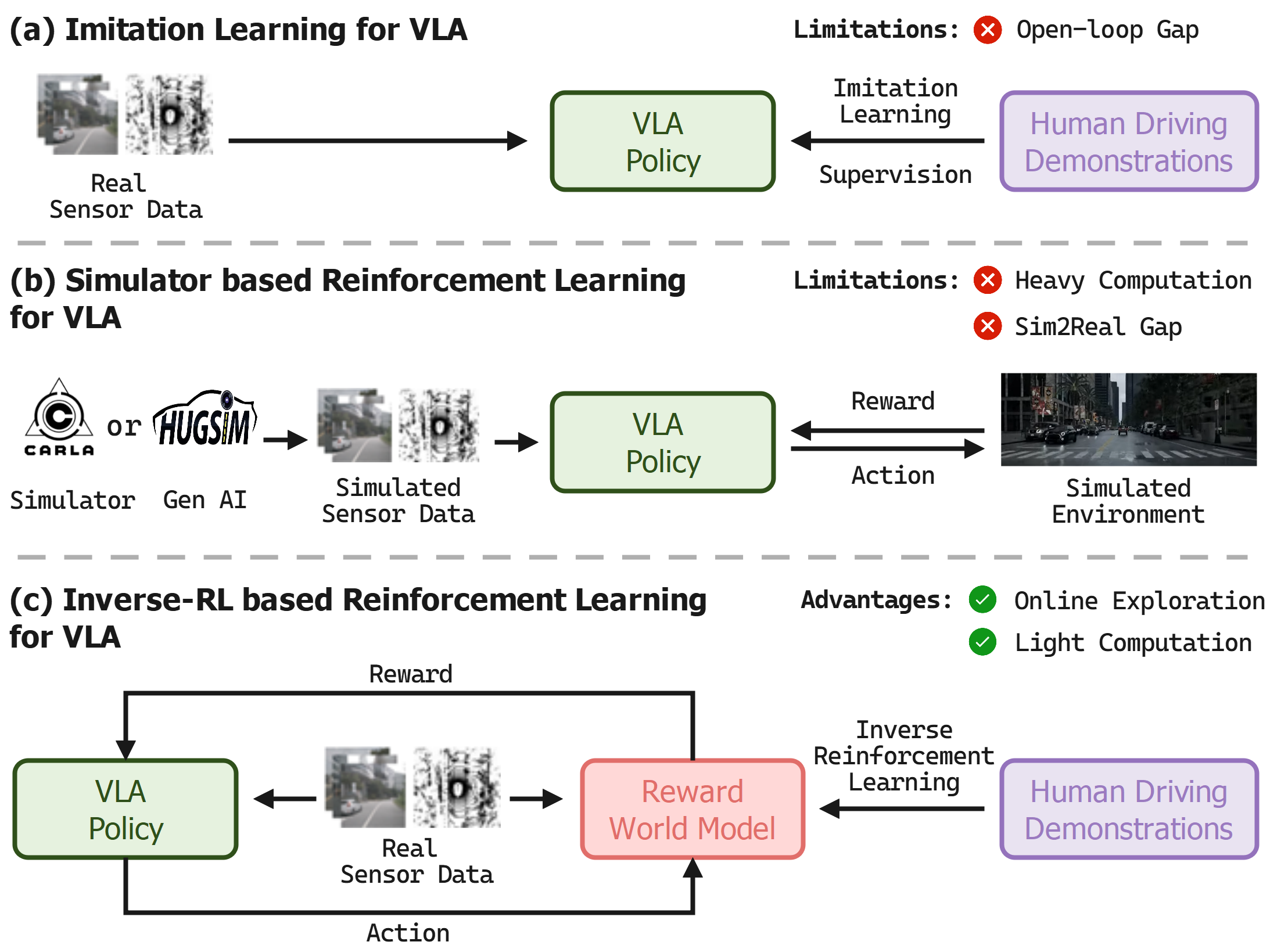} 
    \caption{ \textbf{Different paradigms of VLA autonomous driving (AD)} a).Imitation learning for VLA AD. b). Simulator-based reinforcement learning for VLA AD.  c). IRL-VLA explores improving high-capacity VLA with scalable reinforcement learning without heavy simulator.}
    \label{fig:abs} 
\end{figure}

\vspace{-0.5cm}
\section{Introduction}


End-to-end autonomous driving has emerged as a significant and rapidly growing research area. With the abundance of available human driving demonstrations, there is significant potential to learn human-like driving policies from large-scale datasets.  Methods such as UniAD~\cite{hu2023planning}, VAD~\cite{jiang2023vad} take sensor data as input and directly regress a single-mode trajectory within one fully optimizable model. SparseDrive~\cite{sun2024sparsedrive} further explores the sparse representations and proposes a symmetric sparse perception module with a parallel motion planner. Leveraging diffusion policy in robotics, DiffusionDrive~\cite{liao2025diffusiondrive}, Diffusion Planer~\cite{janner2022planning} and DiffsemanticFusion\cite{sun2025diffsemanticfusionsemanticrasterbev} generate diverse driving actions via an anchored Gaussian distribution design and a carefully designed denoising process. Despite substantial advancements in end-to-end autonomous driving, the system's performance remains vulnerable, exhibiting notable degradation in long-tail driving scenarios. Recent approaches have sought to mitigate this challenge by harnessing the extensive world knowledge embedded in Vision-Language Models (VLMs), namely Vision-Language-Action (VLA) models which take video as input and output driving trajectories and actions directly.

Methods~\cite{jiang2024senna, zhou2025opendrivevla} connect VLM to an end-to-end model to improve trajectory planning accuracy.
RecogDrive~\cite{li2025recogdrive} introduces a novel driving framework which combines a vision-language model with a diffusion-based planner, and simulator-assisted reinforcement learning to generate safe, human-like trajectories.
SimLingo~\cite{renz2025simlingo} introduces an action dreaming task to rigorously assess instruction-conditioned behaviors. ORION~\cite{fu2025orion} integrates vision-language reasoning with generative trajectory planning via a query-based temporal module named QT-Former, and employs a planner based on a variational autoencoder (VAE).

While imitation learning VLA approaches, as shown in Fig.~\ref{fig:abs}.a, achieve superior performance empowered by the remarkable capabilities of VLM, we argue that the full potential of large-scale models remains underexplored due to imitation learning within an open-loop framework which tends to replicate the recorded behaviors in the dataset. This is because driving is inherently a multi-target and multi-modality task, where multi-modality reflects the presence of multiple optimal solutions while multi-target entails satisfying diverse evaluation criteria (e.g., collision avoidance, drivable area compliance, traffic rule compliance, etc.). A more naturalistic strategy involves enabling the model to autonomously explore within a simulated environment, as shown in Fig.~\ref{fig:abs}.b, thereby mimicking the experiential learning process observed in the real world. Nevertheless, the development of a fully interactive and close-loop simulation framework remains a significant technical challenge: 1). Sim2Real domain gap 2). Heavy computational overhead.

In this paper, as shown in Fig.~\ref{fig:abs}.c, we introduce a novel close-loop Reinforcement Learning via \textbf{I}nverse \textbf{R}einforcement \textbf{L}earning framework with a self-built VLA approach, named IRL-VLA. Leveraging our framework, a real-time reward world model (RWM) is designed and learned via inverse reinforcement learning from diverse policy. It captures the multi-modality and multi-target essences of driving meanwhile scalable to large amount of real-world data in a cost-effective way, circumventing Sim2Real domain adatption. We apply the learned RWM to guide reinforcement learning of our VLA model. Our approach achieves state-of-the-art performance in NAVSIM v2 end-to-end driving benchmark, 45.0 EDPMS as 1st runner up in CVPR2025. To the best of our knowledge, our IRL-VLA is the first close-loop VLA approach via end-to-end reinforcement learning including sensor input. The key contributions of our work are summarized as follows:
\vspace{-0.1cm}
\begin{enumerate}[noitemsep]

        \item We propose \textbf{IRL-VLA}, a pioneering framework for reinforcement learning from simulator feedback, specifically designed for Vision-Language-Action (VLA) models. To replace computationally expensive simulator-based reward computation, we introduce an efficient reward world model (RWM) based on inverse reinforcement learning, enabling scalable and effective reward estimation. This learned reward model is then used to train VLA agents via reinforcement learning, significantly enhancing their practicality. To the best of our knowledge, this is the first work to develop a reinforcement learning-based VLA model for autonomous driving without relying on simulator during training.
        \item We propose a novel VLA model that achieves superior performance in both imitation learning and reinforcement learning settings, enabling optimal performance across diverse training paradigms.
        \item Our IRL-VLA framework achieves superior performance on the NAVSIM v2 end-to-end driving benchmark in CVPR2025 challenges. These results demonstrate the effectiveness and generalizability of our approach. 

\end{enumerate}

\section{Related Work}
\label{sec:Related Work}

{\bf End-to-end Autonomous Driving}: Research interest in end-to-end autonomous driving has surged due to its fully differentiable design integrating modular tasks, i.e. perception, prediction, and planning, which enables optimization in pursuit of the ultimate goal. UniAD~\cite{hu2023planning} introduces a comprehensive framework that incorporates full-stack driving tasks within a single network. VAD~\cite{jiang2023vad} represents the driving scene in a fully vectorized manner—encompassing both agent trajectories and map elements—thereby eliminating the need for computationally intensive rasterized representations. Sparsedrive~\cite{sun2024sparsedrive} further explores the sparse presentation and proposes a symmetric sparse perception module and a parallel motion planner. Leveraging diffusion policy in robotics, DiffusionDrive~\cite{liao2025diffusiondrive}, Diffusion Planer~\cite{janner2022planning} and DiffSemanticFusion~\cite{sun2025diffsemanticfusionsemanticrasterbev} generate diverse driving actions via an anchored Gaussian distribution design and appropriate denoising process.

{\bf Vision Language Action Models in Autonomous Driving}: Recent methods such as those proposed by~\cite{jiang2024senna, zhou2025opendrivevla, hwang2024emma} establish bridge between Vision-Language Models (VLMs) and end-to-end frameworks to enhance trajectory planning accuracy.
Recogdrive~\cite{li2025recogdrive} introduces a novel end-to-end driving architecture that combines a vision-language model, a diffusion-based planner, and simulator-assisted reinforcement learning to produce safe and human-like trajectories. SimLingo~\cite{renz2025simlingo} introduces the Action Dreaming task to provide a rigorous evaluation of instruction-conditioned driving behaviors.
Furthermore, ORION~\cite{fu2025orion} proposes the fusion between vision-language reasoning and trajectory planning using QT-Former and VAE. However, these approaches rely on imitation learning, which limits their generalization to real-world multi-modal and multi-target driving scenarios.

{\bf Reinforcement Learning in Autonomous Driving}: Reinforcement Learning (RL) has emerged as a promising approach, with demonstrated success in large language models (LLMs)  and games~\cite{guo2025deepseek, achiam2023gpt}. In the context of autonomous driving, RL has been employed to address specific decision-making challenges and complex driving scenarios. RAD~\cite{gao2025rad} employs reinforcement learning to train an end-to-end autonomous driving agent within a photorealistic 3D Gaussian Splatting (3DGS) simulation framework. However, this method is limited to off-line policy learning due to heavy computation in sensor rendering and Sim2Real domain gap remains unaddressed. Others~\cite{wang2023efficient, zhou2023accelerating} have proposed learning-based trajectory planning frameworks in which actions are represented directly as ego-centric planned trajectories. CarPlanner~\cite{zhang2025carplanner} proposes an RL-based planner surpassing both IL- and rule-based state-of-the-arts (SOTAs) on  the challenging large-scale real-world dataset nuPlan. DiffVLA~\cite{jiang2025diffvla} proposes an efficient VLA model with hierarchical coarse to fine diffusion-based trajectory generation with VLM navigation guidance. Though it achieves state-of-the-art performance on NAVSIMv2 benchmark, its imitation learning set-up constrains its potential. Our IRL-VLA framework extend RL beyond planner to the entire VLA model architecture, which further improves the upper bound of model performance.

\vspace{-0.2cm}
\section{Method}
\begin{figure*}[htbp!] 
    \centering
    \includegraphics[width=1.0\textwidth]{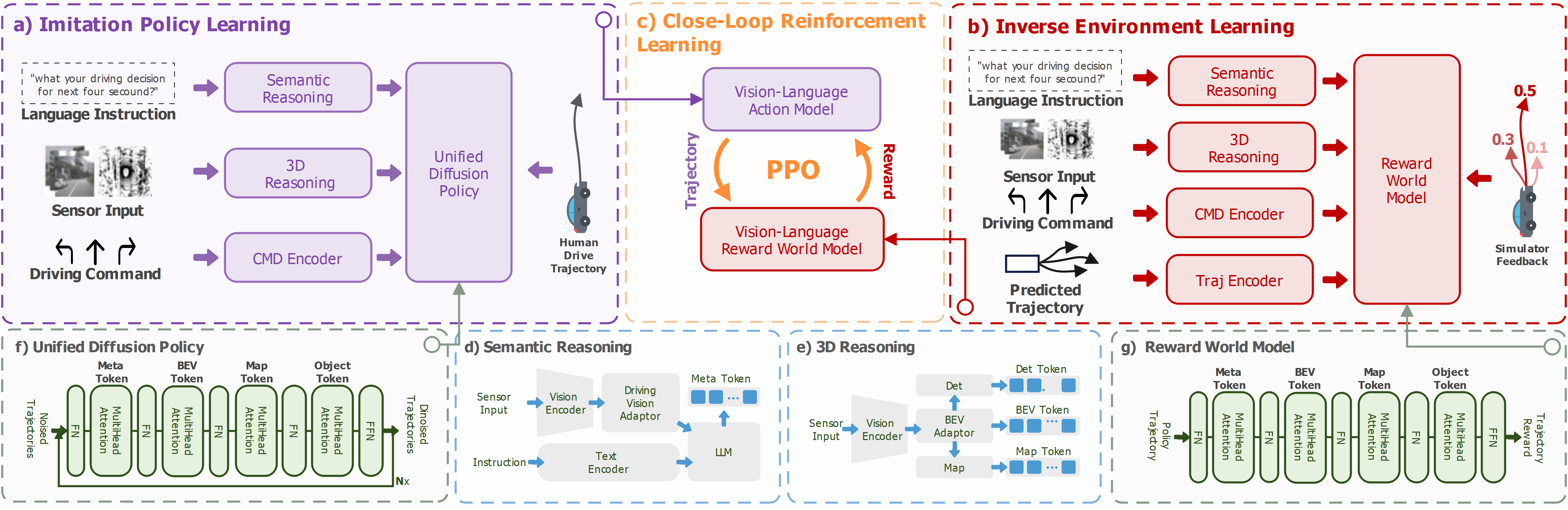} 
    \caption{\textbf{Overview of the IRL-VLA Framework.} This figure illustrates the three-stage pipeline of our close-loop Reinforcement Learning via Reward World Model  framework for Vision-Language-Action (VLA) in autonomous driving. a) Imitation Policy Learning initializes the VLA model as a supervised policy via sensor input and planning trajectories. b) Inverse Environment Learning constructs the Reward World Model (RWM) via pretrained VLA planning trajectories. c)  Close-Loop Reinforcement Learning optimizes the policy using PPO and the RWM. Subfigures (e), (f), and (g) detail for the Unified Diffusion Policy, Semantic Reasoning, and Reward World Model, respectively}
    \vspace{-0.2cm}
    \label{fig:Overview} 
\end{figure*}

In this section, we present the details of our proposed Vision-Language-Action (VLA), trained through close-loop reinforcement learning with a Reward World Model,
as illustrated in Fig.~\ref{fig:Overview}. In Sec.~\ref{sec:immitation policy learning}, we introduce a novel VLA model and pretrain the VLA model using imitation learning to establish a baseline understanding of driving behaviors. In Sec.~\ref{sec:inverse env learning}, we presents a Reward World Model (RWM) via inverse reinforcement learning to generate environment-specific rewards. In Sec.~\ref{sec:RL with RWM}, reinforcement learning environment is constructed where the RWM provides rewards to fine-tune the VLA model.

\subsection{Problem Formulation}
\label{sec:problem formulation}
In autonomous driving, the end-to-end driving outputs the future trajectory $T_{traj}$ or future action $A$ based on sensor input data $S_{sensor}$, like multi-view camera image or lidar/radar point cloud, ego status $S_{ego}$ (ego speed and ego acceleration):
$$ T_{traj} = \pi_{policy}(\theta | {S_{sensor}, S_{ego}})$$
Where the trajectory $T_{traj}$ can be represented as a sequence of vehicle way points (coordinates and heading) $P= \{p_0, p_1, ...,p_l\}$ in the current ego-vehicle coordinate system, where each waypoint $p_i \in \mathbb{R}^3$ is defined as $p_i = (x_i, y_i, \theta_i)$, with $(x_i, y_i)$ denoting the 2D position and $\theta_i$ representing the heading angle. Alternatively, a sequence of agent actions can also be represented as $A = \{a_1, a_2, \ldots, a_l\}$, where each action $a_i$ has the same semantic meaning as $p_i$. The parameter $l$ denotes the prediction horizon, i.e., the number of future steps to be predicted.

\subsection{Imitation Policy Learning}
\label{sec:immitation policy learning}
\textbf{Vision Language Action}: Inspired by recent advancements in dual-process theory within the field of embodied intelligence, we propose an efficient VLA model for autonomous driving which comprises three distinct modules: (1) a semantic reasoning module for deep scene understanding, (2) a 3D reasoning module for accurate geometric inference, and (3) a unified diffusion-based planner to generate diverge driving trajectories.

\textbf{Semantic reasoning}: As shown in Fig.~\ref{fig:Overview}.d, to enable effective processing and fusion of multimodal information in autonomous driving scenarios, we propose the
VLM command guidance module. This module is built upon the Senna-VLM framework, which leverages a multi-image encoding strategy and multi-view prompting mechanisms to achieve efficient and comprehensive scene understanding.

\textbf{3D reasoning}: As illustrated in Fig.~\ref{fig:Overview}.e, we first employ a BEV vision encoder and an adaptor to encode multi-view image to a feature map in BEV space. Then we utilize a group of detection tokens and map tokens to learning vectorized map element and agent motion information from BEV feature space. 

\textbf{Unified Diffusion Planner}:
As shown in Fig.~\ref{fig:Overview}.f, to generate diverse and informative future trajectory distributions, we employ a diffusion-based approach that processes anchor proposal trajectories with Gaussian noise, the conditional diffusion model learns a robust denoising mechanism capable of capturing the inherent multi-modality of future motion. 

To enhance the denoising process, we hierarchically integrate rich scene semantics—such as BEV tokens, map tokens, and detection tokens—into the trajectory generation pipeline. This ensures that the model synthesizes trajectories consistent with environmental constraints. Following the final conditional decoding step, the multi-modal trajectories are reconstructed from the refined trajectory tokens using a lightweight MLP-based regression head. This enables each mode to align with feasible, interpretable, and scene-compliant motion patterns, improving both realism and adherence to physical constraints.

\textbf{Imitation Policy Learning Loss}: Same as other diffusion-based imitation learning training policy, our VLA decoder $\pi_\theta$ takes as input $N_{anchor}$ noisy anchor trajectories $\{\tau^i_k\}^{N_{anchor}}_{k=1}$ and predicts classification scores $\{\hat{s}_k\}^{N_{anchor}}_{k=1}$ and denoised trajectories $\{\hat{\tau}^i_k\}^{N_{anchor}}_{k=1}$, the training objective combines trajectory reconstruction and classification:
$$
L_{IL} = \sum^{N_{anchor}}_{k=1}[y_kL_{rec}(\hat{\tau}_k, \tau_{gt}) + \lambda BCE(\hat{s_k}, y_k)]
$$
where $\lambda$ balances the simple L1 reconstruction loss $L_{rec}$  and
 binary cross-entropy (BCE) classification loss.

\begin{table*}[t]

\scriptsize
\caption{\textbf{Performance on the Navhard Benchmark.}}

\centering
\scalebox{1.1}{
\begin{tabular}{l| c|c |c| c c c c c c c c c |c}

    \toprule
    Method 
    & Img. Resolution
    & Backbone
    & Stage
    & $\text{NC}$
    & $\text{DAC}$
    & $\text{DDC}$
    & $\text{TLC}$
    & $\text{EP}$
    & $\text{TTC}$
    & $\text{LK}$ 
    & $\text{HC}$ 
    & $\text{EC}$ 
    & $\text{EPDMS}$  \\
    \midrule
    
    PDM-Closed~\cite{dauner2023parting}  & GT Perception & - &   \makecell{Stage 1} & \makecell{94.4} & \makecell{98.8} & \makecell{100} & \makecell{99.5} & \makecell{100} & \makecell{93.5} & \makecell{99.3} & \makecell{87.7} & \makecell{36.0 } & 51.3    \\
    \midrule
    \midrule

    CV & * & * &   \makecell{Stage 1} &
    \makecell{88.8} & \makecell{42.8} & \makecell{70.6} & \makecell{99.3} & \makecell{77.5} & \makecell{87.3} & \makecell{78.6} & \makecell{97.1} & \makecell{60.4} & 26.8    \\
    \midrule

    
    LTF~\cite{chitta2022transfuser} & $256\times 1024$ & ResNet34 &   \makecell{Stage 1} &
    \makecell{96.2} & \makecell{79.5} & \makecell{99.1} & 
    \makecell{99.5} & \makecell{84.1} & \makecell{95.1} & 
    \makecell{94.2} & \makecell{97.5} & \makecell{79.1 }& 
    23.1 \\
    \midrule

    DiffuisonDrive\cite{liao2024diffusiondrive}  & $512\times 2048$ & Resnet34 & \makecell{Stage 1} &
    \makecell{95.9} & \makecell{84.0} & \makecell{98.6} & 
    \makecell{99.8} & \makecell{84.4} & \makecell{96.0} & 
    \makecell{95.1} & \makecell{97.6} & \makecell{71.1} &
    63.2 \\
    \midrule
    
    WOTE\cite{wote} & $512\times 2048$ & Resnet34 &  \makecell{Stage 1} &
    \makecell{97.4} & \makecell{88.2} & \makecell{97.8} & 
    \makecell{99.3} & \makecell{82.7} & \makecell{96.4} & 
    \makecell{90.9} & \makecell{97.3} & \makecell{68} &
    66.7 \\
    \midrule

    Hydra-MDP\cite{li2024hydra} & $512\times 2048$ & V2-99&   \makecell{Stage 1}  &
    \makecell{97.6} & \makecell{96.4} & \makecell{99.2} & \makecell{99.3} & \makecell{80.2} & \makecell{96.9} & \makecell{94.9} & \makecell{97.8} & \makecell{58.7} & 73.1   \\
    
    GTRS-Dense\cite{gtrs} & $512\times 2048$ & V2-99&   \makecell{Stage 1}  &
    \makecell{98.7} & \makecell{95.8} & \makecell{99.4} & \makecell{99.3} & \makecell{72.8} & \makecell{98.7} & \makecell{95.1} & \makecell{96.9} & \makecell{40.} & 73.1   \\

    GTRS-Aug\cite{gtrs} &$512\times 2048$ & V2-99&   \makecell{Stage 1 }  &
    \makecell{\textbf{98.9}} & \makecell{\textbf{95.1}} & \makecell{99.2} & \makecell{\textbf{99.6}} & \makecell{76.1} & \makecell{\textbf{99.1}} & \makecell{94.7} & \makecell{\textbf{97.6}} & \makecell{54.2} & 
    74.3 \\
    \midrule

    IRL-VLA-PT(Our) & $256 \times 704$ & V2-99 &  \makecell{Stage 1} &
    \makecell{98.3} & \makecell{92.4} & \makecell{\textbf{99.3}} & 
    \makecell{99.5} & \makecell{83.9} & \makecell{\textbf{97.1}} & 
    \makecell{98.9} & \makecell{\textbf{97.6}}  & \makecell{\textbf{76.0}} &
    74.4 \\

    IRL-VLA-RL(Our) & $256 \times 704$ & V2-99 &  \makecell{Stage 1} &
    \makecell{96.9} & \makecell{91.3} & \makecell{99.1} & \makecell{\textbf{99.6}} & \makecell{96.2} & \makecell{96.2} & \makecell{\textbf{98.0}} & 
    \makecell{97.3}  & \makecell{72.4} &
    \textbf{74.9} \\
    
\bottomrule
\end{tabular}
}

\label{table:navhard}

\end{table*}

\subsection{Inverse Environment Learning}
\label{sec:inverse env learning}
\textbf{Reward Data Collection}: To develop an effective Reward World Model (RWM), a comprehensive dataset is essential. Our approach leverages human-designed metrics from the Ego-Pseudo Driving Metric System (EPDMS)~\cite{cao2025pseudo, dauner2025navsim}, which comprises nine sub-scores: No At-Fault Collision (NC), Drivable Area Compliance (DAC), Driving Direction Compliance (DDC), Traffic Light Compliance (TLC), Ego Progress (EP), Time to Collision (TTC), Lane Keeping (LK), History Comfort (HC), and Extended Comfort (EC), along with a weighted summation score, denoted EPDMS. We exclude EC from our model’s scope, as it requires two separate simulations per scene. These metrics provide detailed insights into the environment and agent interactions. However, imitation data alone often lack diversity, as they do not fully capture varied trajectories across diverse scenarios.

To enhance score and trajectory diversity and ensure model generalization, we employ three strategies. First, We record the trajectories at every time step of the diffusion process and the corresponding EPDM scores. Second, instead of using a fixed trajectory set, we sample multiple trajectory patterns from human demonstration data using K-means clustering, with $K$ ranging from 32 to 8192. Third, we apply multiple ego poses during simulations for each scene in the NAVSIM dataset to generate diverse samples.

\textbf{Reward World Model}: We propose a Reward World Model (RWM) as a lightweight, data-driven alternative to traditional simulators, enabling close-loop evaluation of autonomous systems and collection of downstream driving statistics, such as collision rate, traffic rule compliance, and driving comfort, via inverse reinforcement learning. The RWM eliminates the need for computationally intensive simulators and mitigates the sim-to-real gap by directly modeling the reward structure based on real-world demonstrations. Its architecture, illustrated in Fig.~\ref{fig:Overview}(b), mirrors that of the agent, using multi-perspective camera information and the agent’s predicted future trajectories as inputs. The RWM predicts the future reward of the agent within a simulated environment.

The RWM models the relationship between scores and the environment for a given trajectory using a rule-based simulator. The NAVSIM simulator generates three types of scores. The EP score measures the ego vehicle's progress along the centerline, ranging from $\left[ 0,1 \right]$. The DAC, TLC, TTC, LK, and HC scores are binary, taking values of $\left\{ 0,1 \right\}$, as they assess whether the ego vehicle adheres to predefined driving rules. The NC and DDC scores have values of $\left\{ 0,0.5,1 \right\}$, as they impose fewer penalties when the ego vehicle's behavior is not at fault.
The metrics are modeled as follows:
$$
\hat{r}_{m} = \mathrm{MLP}_{m}({f_{\text{traj}}}), \quad 
m \in 
\begin{cases}
\text{NC}, \text{DAC}, \text{DDC}, \text{TLC}, \\
\text{EP}, \text{TTC}, \text{LK}, \text{HC}
\end{cases}
$$
where $\hat{r}_m$ represents the rewards from different metrics, $f_{\text{traj}}$ represents the trajectory feature amd $m$ stands for different sub-metrics in PDMS. The trajectory feature is extracted from BEV-space features at waypoints along a given trajectory, serving as a hidden representation of the interactions between the trajectory, surrounding agents, and the environment. The final reward $\hat{R}$ is computed as a weighted sum of the individual components:
$$
\hat{r}_{epdms} = \sum_{m} w_{m} \cdot \hat{r}_{m}
$$
where the weights $w_{m}$ for each metric follow the definition of EPDMS in~\cite{cao2025pseudo}.

\textbf{Reward World Model Optimization}: The RWM is trained to minimize the error between predicted and true metric scores. At each training step, a batch of trajectories and their corresponding true scores is sampled to optimize the RWM. The loss function for training the RWM is formulated as follows:
$$
L_{R_{epdms}} = \sum_{i, m }{w}_{m}\left\|\hat{r}_{m}^{i} - r_{m}^{i}\right\| 
$$
where $\hat{r}_m^{i}$ is the predicted score for metric $m\in{\text{NC}, \text{DAC}, \text{DDC}, \text{TLC}, \text{EP}, \text{TTC}, \text{LK}, \text{HC}}$ for the $i$-th trajectory, and $r_m^{i}$ is the corresponding true score from the simulator.

\subsection{Reinforcement Learning with RWM}
\label{sec:RL with RWM}

While imitation learning provides a strong baseline policy, it is inherently limited by biases and incomplete coverage in offline demonstrations. To overcome these limitations, we employ close-loop reinforcement learning with the RWM to fine-tune the VLA policy as depicted in Fig.~\ref{fig:Overview}.c. We adopt the Proximal Policy Optimization (PPO) algorithm due to its stability and sample efficiency—critical properties when training with a learned reward model prone to approximation errors.

\textbf{Policy Optimization}: The policy optimization process involves iteratively sampling trajectories from the VLA policy, evaluating them via the RWM, and updating the policy parameters to maximize expected cumulative rewards. By providing real-time reward feedback, the RWM eliminates the need for computationally expensive sensor rendering and physics-based simulations. This enables scalable and efficient training, allowing the VLA model to explore diverse driving scenarios and optimize multi-target objectives (safety, efficiency, traffic rule compliance). We use PPO to train the policy $\pi_\theta$ with the RWM, chosen for its stability and sample efficiency when interacting with a learned environment that may introduce approximation errors. The optimization process follows these steps:

\begin{algorithm}[h]
\caption{Policy Optimization with PPO in RWM}
\label{alg:ppo}
\begin{algorithmic}[1]
\State \textbf{Input}: Trained RWM \(\hat{\mathcal{R}}\), policy \(\pi_\theta\), value function \(V_\phi\), learning rate \(\eta\), clipping parameter \(\epsilon\), number of epochs \(K\), trajectory length \(l\),
\For{each iteration}
    \State Initialize empty trajectory set \(\mathcal{T}\)
    \For{\(t = 1\) to \(T\)}
        \State Sample \(\mathbf{a}_t \sim \pi_\theta(\cdot | \mathbf{s}_t)\)
        \State Compute \(\hat{\mathbf{s}}_{t+1} \sim \hat{\mathcal{T}}(\mathbf{s}_t, \mathbf{a}_t)\), \(\hat{r}_{t+1} = \hat{\mathcal{R}}(\mathbf{s}_t, \mathbf{a}_t)\)
        \State Store \((\mathbf{s}_t, \mathbf{a}_t, \hat{\mathbf{s}}_{t+1}, \hat{r}_{t+1})\) in \(\mathcal{T}\)
        \State Set \(\mathbf{s}_{t+1} = \hat{\mathbf{s}}_{t+1}\)
    \EndFor
    \State Compute advantages \(A_t\) using GAE
    \For{\(k = 1\) to \(K\)}
        \State Compute policy loss:
        \State \hspace{1em} \(L^{\text{CLIP}}(\theta)\)
        \State Compute value loss:
        \State \hspace{1em} \(L^{\text{VF}}(\phi) = \mathbb{E}_t \left[ (V_\phi(\mathbf{s}_t) - R_t)^2 \right]\)
        \State Update \(\theta \gets \theta + \eta \nabla_\theta L^{\text{CLIP}}(\theta)\)
        \State Update \(\phi \gets \phi - \eta \nabla_\phi L^{\text{VF}}(\phi)\)
    \EndFor
\EndFor
\State \textbf{Output}: Optimized policy \(\pi_\theta\)
\vspace{-0.1cm}
\end{algorithmic}
\end{algorithm}

Inspired by diffusion-based planning methods~\cite{ren2024diffusion, gtrs}, our diffusion policy $\pi_\theta$ can be viewed as an intrinsic Markov decision process that starts from Gaussian noise and progressively denoises it to produce a sequence of actions. Following this paradigm, we generate a set of trajectories $T_{trajs}$ and record their complete diffusion processes. For a single trajectory, the diffusion progression is defined as

\begin{equation}
\mathbf{T_{traj}} = \bigl(T_{traj\tau}, T_{traj\tau-1}, \dots, T_{traj0}\bigr),
\end{equation}
where $\tau$ denotes the total number of denoising steps.

In our framework, the RWM evaluates each trajectory using a multi-criteria scoring system, combining comfort metrics (EP, LK, HC) with safety metrics (NC, DAC, DDC, TLC, TTC). These are aggregated into an EPDMS-based score $r_{epdms}$. This score is combined with the value estimation from a critical network, and the advantage values are computed using GAE~\cite{schulman2018highdimensionalcontinuouscontrolusing} as the $R_{epdms}=GAE(r_{epdms}+v_{epdms})$. Further, inspired by~\cite{shao2024deepseekmathpushinglimitsmathematical, li2025recogdrive}, to improve stability across the batch, we adopt the group-standardized advantage the subsequent processing step involves derivation of group-standardized advantage values,
\begin{equation}
\hat A_{i} \;=\; \frac{R_{epdms} - \mathrm{mean}\bigl(R_{1...T_{trajs}}\bigr)}%
{\sqrt{\mathrm{var}\bigl(R_{1..T_{trajs}}\bigr)}},
\quad
i=1,\dots,T_{trajs}.
\end{equation}

In diffusion progress, each conditional transition is modeled by a Gaussian policy \cite{black2024trainingdiffusionmodelsreinforcement}:
\begin{equation}
\pi_\theta\left(x_{t-1} \mid x_t, c\right)
= \mathcal{N}\bigl(x_{t-1};,\mu_\theta(x_t,c,t),\sigma_t^2 I\bigr),
\end{equation}
where \(\mu_\theta(x_t,t,c)\) is the model‐predicted mean and \(\sigma_t^2 I\) is the fixed covariance, $c$ include current vehicle
states, semantic reasoning features, 3D reasoning features, and navigation command.

Inspired by recogdrive~\cite{li2025recogdrive}, The joint log-likelihood of the entire trajectory under this diffusion policy can therefore be expressed as
\begin{equation}
\log \pi_\theta\bigl(\mathbf{x}_{0:T}\bigr)
= \sum_{t=1}^\tau \log \pi_\theta\bigl(x_{t-1}\mid x_t, c\bigr).
\qquad
\end{equation}
For policy optimization, we adopt the reinforcement learning via PPO algorithm. The reinforcement learning loss is formulated as follows:

\begin{align}
L_{RL} &=
-\frac{1}{T_{trajs}}\sum_{i=1}^T\frac{1}{\tau}\sum_{t=1}^\tau
\gamma^{\,t-1}\,\log \pi_\theta\bigl(x_{t-1}^{(i)}\mid x_t^{(i)}, c\bigr)\,\hat A_{i} \nonumber \\
&\quad - D_{KL}\!\left[\pi_{\theta} \,\big\|\, \pi_{ref}\right] 
\end{align}

where $\gamma$ is the discount coefficient (mitigating instability in early denoising steps),  and $ x_{t-1}, x_t$ are sampled from the reference policy $\pi_{\mathrm{RL}}$, and $D_{KL}\!\left[\pi_{\theta} \,\big\|\, \pi_{ref}\right] $ is the KL divergence between the reference policy $\pi_{ref}$ and current policy $\pi_{\theta}$.

By leveraging RWM-assisted reinforcement learning, the diffusion-based planner acquires the capability to generate safe and comfortable trajectories through active exploration, moving beyond simple imitation and introducing cognitive reasoning into the framework. The final policy optimization loss combines the reinforcement learning objective with a behavior cloning term to maintain stability and prevent catastrophic forgetting of the pretrained policy:

\begin{equation}
L = L_{\text{RL}} + w_{IL}L_{\text{IL}}
\end{equation}

Which $w_{IL}$ is the behavior cloning loss weight. This combined loss ensures stable, effective policy optimization, leveraging the RWM to guide the VLA model toward optimal driving behaviors.

\section{Implementation Details}
\label{sec:imp details}
The IRL-VLA model is implemented with a V2-99 backbone and processes multi-view camera inputs at a resolution of 256 × 704. The imitation learning stage (IRL-VLA-PT) is pre-trained for 100 epochs using the AdamW optimizer with a learning rate of $10^{-4}$ and a batch size of 32. The Reward World Model is trained using inverse reinforcement learning with binary cross entropy loss for those metrics in EPDMS in range of $\left\{ 0,1 \right\}$, mean square error loss for the metric in range of $\left[ 0,1 \right]$ and cross entropy loss for those metrics in range of $\left\{ 0,0.5,1 \right\}$, leveraging expert demonstrations and simulator feedbacks. For the reinforcement learning stage (IRL-VLA-RL), we employ Proximal Policy Optimization (PPO) with a clipping parameter $\epsilon=0.2$, a discount factor $\gamma=0.99$, and a generalized advantage estimation (GAE) parameter $\lambda = 0.95$. Training is conducted on 8 NVIDIA A100 GPUs.

\section{Experiments}
\label{sec:Experiments}
 In our experiments, we focus on the following questions:
 \vspace{-0.2cm}
\begin{enumerate}[noitemsep]
\item How does IRL-VLA perform on commonly open-loop and close-loop autonomous driving benchmarks?
\item How do the proposed techniques and implementation details impact IRL-VLA performance?
\end{enumerate}

\subsection{Experimental Settings}
\label{sec:Experiments Settings}

\textbf{Dataset and Metrics.} NAVSIM is a planning-oriented autonomous driving dataset built on OpenScene,
a redistribution of nuPlan. It provides eight 1920×1080 cameras and a fused LiDAR point cloud aggregated from five sensors across the current and three previous frames. The dataset is split into
navtrain (1,192 training scenes) and navhard (136 evaluation scenes).

The NAVSIM benchmark provides a non-reactive simulation environment and employs the Extend Predictive Driver Model Score (EPDMS) as its close-loop planning metric:
\begin{multline}
\text{EPDMS} = 
\underbrace{
\prod_{m \in \mathcal{M}_\text{pen}} \text{filter}_m(\text{agent}, \text{human})
}_{\text{penalty terms}}
\cdot \\
\underbrace{
\frac{ \sum_{m \in \mathcal{M}_\text{avg}} w_m \cdot \text{filter}_m(\text{agent}, \text{human}) }
     { \sum_{m \in \mathcal{M}_\text{avg}} w_m }
}_{\text{weighted average terms}}
\label{eq:epdms}
\end{multline}
where EPDMS integrates two sub-metrics group:  $\mathcal{M}_\text{pen} = \{\text{NC}, \text{DAC}, \text{DDC}, \text{TLC}\}$ and $\mathcal{M}_\text{avg} = \{\text{TTC}, \text{EP}, \text{HC}, \text{LK}, \text{EC}\}$. No At-Fault Collision (NC), Drivable Area Compliance (DAC), Driving Direction Compl (DDC), Lane Keeping(LK),Time-to-Collision (TTC), History Comfort (HC), Extended Comfort(EC), Traffic Light Compl. (TLC) and Ego Progress (EP) to produce a comprehensive close-loop planning score.

\begin{table*}[t]
  \centering
  \small
\caption{\textbf{Ablation study on the proposed components of our proposal hierarchical reasoning diffusion VLA agent.} We evaluate the effect of driving 3D reasoning, semantic reasoning, diffusion planner for reasoning diffusion VLA agent on NAVSIM navhard-real evaluation.}
  \setlength{\tabcolsep}{3pt}
  \renewcommand{\arraystretch}{0.9}
  \scalebox{1.1}{ 
  \begin{tabular}{
    c
    c c c c c c c c
    c c c c c c c c c | c
  }
    \toprule
    ID 
      & \makecell{3D \\ Reasoning} 
      & \makecell{Semantic \\ Reasoning} 
      & \makecell{Diffusion \\ Planner} 
      & NC  
      & DAC 
      & DDC 
      & TLC
      & LK
      & TTC 
      & EP  
      & HC 
      & EC
      & \cellcolor{gray!30} EPDMS \\
    \midrule
    1 
      & \ding{51} & \ding{55} & \ding{55} 
      & 98.4& 89.6 & 99.4 & 99.6 & 96.2 & 96.4 & 81.1 & 96.4 & 70.2 &\cellcolor{gray!30} 70.0\\
    2 
      & \ding{51} & \ding{51} & \ding{55} 
      & 98.2 & 89.3 & 99.4 & 99.6 & 97.3 & 96.0 & 83.5 & 97.5 & 79.5 & \cellcolor{gray!30} 71.4 \\
    3 
      & \ding{51} & \ding{51} & \ding{51} 
      & 98.3 & 92.4 & 99.3 & 99.5 & 99.6 & 97.1 & 83.9 &  97.6 & 76.0 & \cellcolor{gray!30}  74.4\\
    \bottomrule
  \end{tabular}}
  \label{tab:abl_1}
\end{table*}

\subsection{Comparison with State-of-the-arts}

Table.\ref{table:navhard} presents the performance of IRL-VLA compared to baseline methods on the Navhard benchmark. Our pretrained model (IRL-VLA-PT) achieves competitive results across multiple metrics, with an EPDMS of 74.4, outperforming several state-of-the-art methods such as DiffusionDrive (63.2), WOTE (66.7), and GTRS-Aug (74.3).  Compared to scorer-based models like GTRS-Dense and GTRS-Aug, which leverage scoring mechanisms to enhance safety metrics such as No Collision (NC,98.9 for GTRS-Aug) at the expense of comfort-related scores like Extended Comfort (EC, 54.2 for GTRS-Aug), our IRL-VLA-PT model achieves significant improvements in EP (83.9 vs. 76.1) and EC (76.0
vs. 54.2) while maintaining near-comparable safety performance (NC: 98.3 vs. 98.9). This balance underscores the effectiveness of our VLA architecture in optimizing both safety and comfort without relying on explicit scoring mechanisms.

\begin{table*}[htbp]
    \centering
    \caption{\textbf{Ablation study on the impact of different imitation learning loss weights}}
    \begin{tabular}{c|ccccccccc|c}
        \toprule
        $w_{IL}$  & NC & DAC & DDC & TLC & EP & TTC & LK & HC & EC & \cellcolor{gray!30}EPDMS \\
        \midrule
        1.0 & 97.0 & 91.1 & 97.4 & 99.6 & 96.7 & 94.9 & 97.1 & 97.1 & 69.3 & \cellcolor{gray!30}73.9 \\
        
        0.5 & 96.9 & 91.3 & 99.1 & 99.6 & 96.2 & 96.2 & 98.0 & 97.3 & 72.4 & \cellcolor{gray!30}74.9 \\
        
        0.1 & 96.7 & 90.2 & 98.1 & 99.3 & 97.3 & 96.0 & 97.6 & 97.1 & 69.8 & \cellcolor{gray!30} 73.4\\
        \bottomrule
    \end{tabular}
    \label{tab:abl_2}
\end{table*}

\subsection{Ablation Studies}
\label{sec:Ablation Studies}
To evaluate how the proposed techniques and implementation details impact IRL-VLA performance, we conduct two ablation studies. These studies examine the best VLA structure, the effect of the reward world model, and the importance of combining RL and IL. 

\textbf{Ablation study on hierarchical reasoning diffusion VLA agent} Tab.\ref{tab:abl_1} presents an ablation study on our proposed hierarchical reasoning diffusion VLA agent of IRL-VLA. When training via human driving demonstrations data with 3d reasoning only, the model achieves a EPDMS of 70.0 on Navhard-real benchmark. Adapting the semantic reasoning with our high level driving command query increases EPDMS by 1.4. Finally, introducing the diffusion planner for continuous trajectory prediction further achieves a EPDMS by 74.4 with 3.0 improvement. Demonstrating the value of our hierarchical reasoning diffusion VLA scheme a strong pretrain performance in producing ucing safer and more comfortable driving behavior.

\textbf{Ablation study on Imitation Loss Weight}: Tab.\ref{tab:abl_2} examines the impact of the imitation loss weight $w_{IL}$. When $w_{IL}=1.0$, imitation learning contributes equally with the reinforcement learning. When $w_{IL}=0.1$ the imitation learning term will weaken leading to the collapse of the training. Finally, setting $w_{IL}=0.5$ achieved the best trade-off between imitation learning and reinforcement learning, which yields the highest EPDMS by 74.9.

\section{Conclusions}
\label{sec:Conclusions}

In this paper, we introduced IRL-VLA, a novel close-loop Reinforcement Learning via Reward World Model framework for Vision-Language-Action (VLA) models in end-to-end autonomous driving. Our three-stage approach—imitation policy learning, inverse environment learning, and close-loop reinforcement learning—addresses the limitations of open-loop imitation learning and simulator-based training. By pretraining a VLA model with semantic and 3D reasoning modules and a diffusion-based planner, constructing a lightweight Reward World Model (RWM) via inverse reinforcement learning, and fine-tuning the policy using PPO, IRL-VLA achieves state-of-the-art performance on the NAVSIM v2 CVPR challenge benchmark, scoring 45.0 EDPM5 and securing the 1st runner-up position in the CVPR 2025 Autonomous Grand Challenge. And also show state of the art performance, scoring 74.9, in NAVSIM Navhard real benchmark. To our knowledge, IRL-VLA is the first close-loop VLA approach incorporating sensor inputs without relying on simulators. Our contributions include a pioneering RL framework for VLA models, an efficient RWM for scalable reward computation, and demonstrated generalizability, paving the way for future advancements in close-loop autonomous driving.












\clearpage
\bibliographystyle{ieee_fullname}
\bibliography{egbib}

\end{document}